\documentclass{article}

\usepackage{arxiv}

\usepackage[utf8]{inputenc} 
\usepackage[T1]{fontenc}    
\usepackage{hyperref}       
\usepackage{url}            
\usepackage{booktabs}       
\usepackage{amsfonts}       
\usepackage{nicefrac}       
\usepackage{microtype}      
\usepackage{lipsum}		
\usepackage{graphicx}
\usepackage[numbers]{natbib}
\usepackage{doi}
\usepackage{todonotes}
\usepackage{amsmath}
\usepackage{subcaption}
\usepackage{float}
\usepackage{adjustbox}

\title{Intercepting Unauthorized Aerial Robots in Controlled Airspace Using Reinforcement Learning}

\date{} 					

\author{Francisco Giral\thanks{Corresponding author} \\
	ETSIAE-UPM-School of Aeronautics \& Space\\
	Universidad Politécnica de Madrid\\
	Madrid, Spain \\
	\texttt{fa.giral@alumnos.upm.es} \\
 	\And
	Ignacio Gomez \\
	ETSIAE-UPM-School of Aeronautics \& Space\\
	Universidad Politécnica de Madrid\\
	Madrid, Spain \\
	\texttt{ignacio.gomez@upm.es} \\
  	\And
	Soledad Le Clainche \\
	ETSIAE-UPM-School of Aeronautics \& Space\\
	Universidad Politécnica de Madrid\\
	Madrid, Spain \\
	\texttt{soledad.leclainche@upm.es} \\
}

\renewcommand{\undertitle}{}
\renewcommand{\shorttitle}{Intercepting Unauthorized Aerial Robots with RL}

\hypersetup{
pdftitle={Intercepting Unauthorized Aerial Robots in Controlled Airspace Using Reinforcement Learning},
pdfsubject={Reinforcement Learning, Flight Control, Machine Learning, Aerial Robotics},
pdfauthor={Francisco Giral, Ignacio Gomez, Soledad Le Clainche},
pdfkeywords={Reinforcement Learning, Flight control, Aerial robotics, Dreamer},
}

\begin{document}
\maketitle

\begin{abstract}
The proliferation of unmanned aerial vehicles (UAVs) in controlled airspace presents significant risks, including potential collisions, disruptions to air traffic, and security threats. Ensuring the safe and efficient operation of airspace, particularly in urban environments and near critical infrastructure, necessitates effective methods to intercept unauthorized or non-cooperative UAVs. This work addresses the critical need for robust, adaptive systems capable of managing such threats through the use of Reinforcement Learning (RL). We present a novel approach utilizing RL to train fixed-wing UAV pursuer agents for intercepting dynamic evader targets. Our methodology explores both model-based and model-free RL algorithms, specifically DreamerV3, Truncated Quantile Critics (TQC), and Soft Actor-Critic (SAC). The training and evaluation of these algorithms were conducted under diverse scenarios, including unseen evasion strategies and environmental perturbations. Our approach leverages high-fidelity flight dynamics simulations to create realistic training environments. This research underscores the importance of developing intelligent, adaptive control systems for UAV interception, significantly contributing to the advancement of secure and efficient airspace management. It demonstrates the potential of RL to train systems capable of autonomously achieving these critical tasks.
\end{abstract}

\keywords{Reinforcement Learning \and Flight Control \and Aerial Robotics \and Motion Planning \and Autonomous Systems}


\section{Introduction}

Autonomous vehicles and robotics have gained significant attention in recent years due to their potential to address various contemporary issues. Specifically, aerial robotics and Unmanned Aerial Vehicles (UAVs) are utilized in both civil and military applications, including search and rescue, surveillance and reconnaissance, boundary control, critical installations protection, and autonomous transportation, among others. Currently, these applications face several limitations because existing methods do not generalize well or adapt to environmental changes, thereby hindering full autonomy.

Machine Learning, and particularly Reinforcement Learning (RL), has proven effective at overcoming these limitations. RL has been successfully applied in various fields such as video games \cite{vinyals2019grandmaster, schrittwieser2020mastering}, mathematical discoveries \cite{fawzi2022discovering}, and robotics \cite{haarnoja2024learning, margolis2024rapid}. In robotics and autonomous vehicles, its potential is particularly promising, providing capabilities for embodied systems to make intelligent decisions despite environmental uncertainties or high-dimensional observations that are challenging for traditional methods. In the domain of autonomous aerial vehicles, Reinforcement Learning has achieved state-of-the-art results in tasks such as drone racing competitions \cite{kaufmann2023champion}, autonomous dogfighting \cite{zhu2024mastering, pope2022hierarchical}, and robust control of multiple suspended loads \cite{belkhale2021model}.

The proliferation of aerial vehicles, both manned and unmanned, has led to an increased presence of flying robots in controlled airspace, such as the vicinity of airports. Unauthorized UAVs in these areas pose significant risks, including potential accidents and disruptions to air traffic operations, which can result in financial losses, flight delays, or complete operational shutdowns. Looking ahead to urban air mobility, where an Unmanned Traffic Management (UTM) system will enable piloted and autonomous aerial vehicles to coexist in urban airspace, controlling both authorized and unauthorized UAVs is crucial for maintaining operational safety.

Addressing the need to capture or intercept unauthorized UAVs in specific airspace zones, whether urban or natural, is essential for ensuring the safety of aerial operations and air traffic. One effective solution involves deploying an intelligent system capable of detecting and capturing unauthorized UAVs, which requires robustness and adaptability to varying target movements and maneuvers. Autonomous aerial robots are well-suited for this task.

This research aims to tackle the challenge of intercepting a non-cooperative dynamic target (a UAV) with a pursuing agent. We employ Reinforcement Learning to train a pursuer agent with the objective of intercepting the target swiftly. Previous studies have utilized RL for similar tasks; some have approached the problem using Multi-Agent Reinforcement Learning (MARL), training both pursuers and evaders in adversarial settings with multiple quadcopters in complex environments \cite{marl1, marl2, marl3}. Other works have focused on training the evader to escape attackers \cite{escape1, escape2}. In single-agent RL settings, researchers have trained quadcopters to follow targets using visual sensors and discrete control actions \cite{single1} or decomposed the problem to manage system dynamics at a low level \cite{single2}. Applications have extended to fixed-wing UAVs, intercepting moving targets with fixed trajectories \cite{single3} or training RL agents to follow control setpoints for pursuing targets \cite{single4, de2023deep}.

This work presents a training framework for a fixed-wing UAV pursuer agent using Single-Agent Reinforcement Learning, with the goal of capturing a dynamic evader. The objective is to train a pursuer UAV capable of handling various evading strategies, beyond those encountered during training. We propose a training setup where the pursuer must intercept a dynamic UAV target that employs different evasion strategies, tested by manually controlling the evader. All simulations are conducted using JSBSim \cite{jsbsim}, a high-fidelity Flight Dynamics Model (6 DoF), ensuring both pursuer and evader share the same physics model. Unlike most previous works, we utilize a model-based RL algorithm, DreamerV3 \cite{dreamer}, which leverages imagined trajectories in a world model to train the control policy in a continuous action space, aiming for robust task representation. We compare Dreamer's performance against two well-known model-free RL algorithms, Truncated Quantile Critics (TQC) \cite{kuznetsov2020controlling} and Soft-Actor-Critic (SAC) \cite{sac}, both of which have shown excellent results in continuous control problems. The robustness and generalization capabilities of the trained agents are evaluated under various scenarios, including wind disturbances and sensor noise.

The contributions and advantages of this paper are as follows:

\begin{itemize}
\item We propose a Reinforcement Learning-based training framework to design a pursuer agent in a single-agent setup, capable of intercepting an evader regardless of its strategy. This approach aims to create a pursuer capable of generalizing to different evading strategies, avoiding the complexity of multi-agent RL setups, and allowing exploration of various algorithmic approaches without relying on low-level or setpoint tracking controllers.
\item We compare model-based and model-free RL approaches. Unlike many other studies that focus solely on model-free algorithms, we evaluate the performance of both methods.
\item We validate the system's robustness by introducing sensor noise and wind gust perturbations post-training, exposing the agent to previously unseen scenarios.
\end{itemize}

The paper is organized as follows: Section \ref{methodology} summarizes the methodology, defining the problem, describing the simulation environment, and detailing the reinforcement learning training framework, including the Markov Decision Process (MDP) definition, reward design, and training parameters. Section \ref{results} presents the training and validation results under various scenarios. Finally, we conclude with our findings and future directions.

\section{Methodology} \label{methodology}

This section introduces the proposed method. First, we provide a brief introduction to the problem we aim to address. Next, we detail the simulation environment and the Flight Dynamics Model used to test our methodology. Following that, we describe the algorithms used in this work, with a particular focus on the model-based RL algorithm, DreamerV3. Finally, we explain the modeling of the problem as a Markov Decision Process (MDP), focusing on the observation space, action space, reward definition, and the training framework.

\subsection{Problem Formulation}

Ensuring the safety and efficiency of airspace operations, especially in controlled environments like airports, necessitates addressing the challenge posed by unauthorized or non-cooperative UAVs. These UAVs can disrupt air traffic, cause accidents, and compromise security. Therefore, developing an intelligent system capable of intercepting dynamic flying targets is crucial for maintaining operational integrity and safety. This problem is particularly significant in the context of advancing urban air mobility, where both piloted and autonomous aerial vehicles need to coexist safely.

The goal is to develop an intelligent system capable of intercepting a dynamic flying target as quickly as possible, anticipating the target's movements regardless of its evading actions. We refer to the agent as the pursuer and the dynamic target as the evader. Our objective is to train the pursuer to complete this task using Reinforcement Learning.

The result of training the agent is the trained pursuer's policy, $\pi_{\theta}(a_t| s_t)$, which serves as a controller, mapping the dynamics of the pursuer and the position of the evader to the optimal actions for the control surfaces and motors.

\begin{figure}[ht!]
    \centering
    \includegraphics[width=0.45\textwidth]{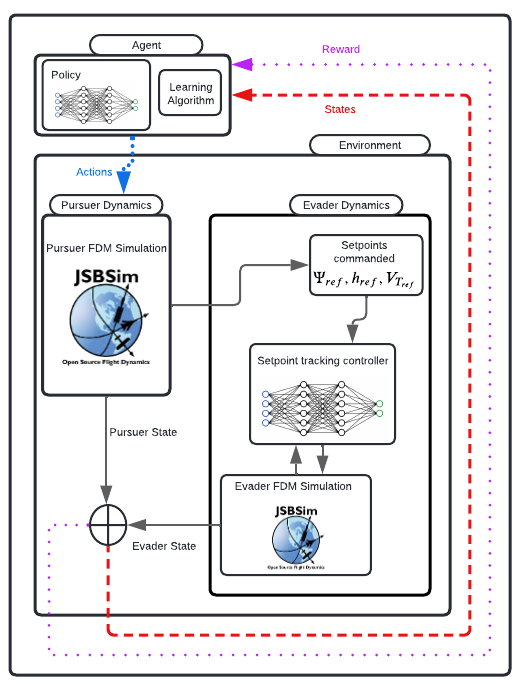}
    \caption{Diagram illustrating the problem setup and the Reinforcement Learning (RL) training framework. The pursuer agent, controlled by the RL policy, interacts with the environment that simulates the flight dynamics of both the pursuer and the evader UAVs. The evader is controlled by a setpoint tracking controller, enabling various evasion strategies. Actions flow are displayed in blue, States and Reward flows are displayed in red and purple respectively, showing the interaction between agent and environment.}
    \label{fig:scheme}
\end{figure}

Figure \ref{fig:scheme} shows the general diagram of the problem and RL training framework, where the agent consists of the controller policy and the training RL algorithm, and the environment encompasses the rest of the problem dynamics. In the environment, two Flight Dynamics Model simulations are carried out for the pursuer and evader air vehicles, respectively.

The evader simulation is independent of the pursuer simulation and is controlled by a setpoint tracking controller designed to track values in heading, altitude, and airspeed ($\Psi_{ref}$, $h_{ref}$, ${V_T}_{ref}$) so that the aircraft can be fully controlled. This controller is a previously trained policy using Reinforcement Learning as well.

During training, these three reference values are used to define the evading strategies for the evader aircraft. Reference values can depend on the pursuer's position, be chosen randomly, or be controlled by a human user in real time. This way, different strategies are composed for training and validation.

Actions taken by the agent modify the pursuer's flight dynamics simulation. A combination of the pursuer's dynamics and the evader's position serves as the observation for the agent. The reward function is defined based on the relative dynamics between the two aircraft. The agent's ultimate goal is to maximize the reward by taking the optimal actions based on the observed state at each time step.

A more detailed explanation of the Markov Decision Process (MDP) formulation is provided in Section \ref{mdp}, delving into the definitions of the state space, action space, reward function, and training framework, including the design of the evasion strategies for the evader.

\subsection{Simulation Environment} \label{section:simulation_env}

We utilize the JSBSim Flight Dynamics Model (FDM) as our physics simulator. JSBSim is an open-source software written in C++ and is considered a high-fidelity simulator for the dynamics of fixed-wing air vehicles.

JSBSim simulates the dynamics of an aircraft using a non-linear flight dynamics model based on the differential system of the form $\dot{x} = f(x, u)$. The equations governing this system are shown in Equations \ref{system1} to \ref{system4}, where \( m \) is the mass of the aircraft, and \( I \) is the moment of inertia around each axis. \( X \), \( Y \), and \( Z \) are the total forces on the \( x \)-axis, \( y \)-axis, and \( z \)-axis of the plane, respectively, expressed in the body frame. \( L \), \( M \), and \( N \) represent moments similarly.

Solving this system of equations allows the calculation of the aircraft's state, given by its position \((x_e, y_e, z_e)\), velocity \((u, v, w)\), Euler angles \((\phi, \theta, \psi)\), and angular rates \((p, q, r)\). For this purpose, JSBSim uses numerical integrators such as Runge-Kutta or Adams-Bashforth.

\begin{align}
\begin{bmatrix} \label{system1}
\dot{x}_e \\
\dot{y}_e \\
\dot{z}_e
\end{bmatrix}
&= [T_{be}(\phi, \theta, \psi)]
\begin{bmatrix}
u \\
v \\
w
\end{bmatrix} \\
\begin{bmatrix} \label{system2}
\dot{\phi} \\
\dot{\theta} \\
\dot{\psi}
\end{bmatrix}
&=
\begin{bmatrix}
1 & \sin\phi \tan\theta & \cos\phi \tan\theta \\
0 & \cos\phi & -\sin\phi \\
0 & \sin\phi / \cos\theta & \cos\phi / \cos\theta
\end{bmatrix}
\begin{bmatrix}
p \\
q \\
r
\end{bmatrix} \\
\begin{bmatrix} \label{system3}
\dot{u} \\
\dot{v} \\
\dot{w}
\end{bmatrix}
&= \frac{1}{m}
\begin{bmatrix}
X \\
Y \\
Z
\end{bmatrix}
+
\begin{bmatrix}
-g \sin\theta \\
g \sin\phi \cos\theta \\
g \cos\phi \cos\theta
\end{bmatrix}
-
\begin{bmatrix}
p \\
q \\
r
\end{bmatrix}
\times
\begin{bmatrix}
u \\
v \\
w
\end{bmatrix} \\
\begin{bmatrix} \label{system4}
\dot{p} \\
\dot{q} \\
\dot{r}
\end{bmatrix}
&=
\begin{bmatrix}
I_{xx} & I_{xy} & I_{xz} \\
I_{yx} & I_{yy} & I_{yz} \\
I_{zx} & I_{zy} & I_{zz}
\end{bmatrix}^{-1}
\left(
\begin{bmatrix}
L \\
M \\
N
\end{bmatrix}
-
\begin{bmatrix}
p \\
q \\
r
\end{bmatrix}
\times
\begin{bmatrix}
I_{xx}p + I_{xy}q + I_{xz}r \\
I_{yx}p + I_{yy}q + I_{yz}r \\
I_{zx}p + I_{zy}q + I_{zz}r
\end{bmatrix}
\right)
\end{align}

In addition to these fixed-wing aircraft equations, JSBSim also models atmospheric disturbances, ground reactions, geodetic modeling, and rotational effects, among others, to enhance simulation realism. Of particular relevance to this work is the turbulence/gusts model used by JSBSim, the Dryden Gusts or Dryden Wind Turbulence model. This model is commonly used in aeronautics due to its suitability for representing atmospheric turbulence effects on aircraft.

The Dryden model equations are given by:

\begin{equation}
\Phi_u(\omega) = \frac{\sigma_u^2 L_u}{\pi} \frac{1}{1 + (L_u \omega)^2}
\end{equation}

\begin{equation}
\Phi_v(\omega) = \frac{\sigma_v^2 L_v}{\pi} \frac{1 + 3(L_v \omega)^2}{[1 + (L_v \omega)^2]^2}
\end{equation}

\begin{equation}
\Phi_w(\omega) = \frac{\sigma_w^2 L_w}{\pi} \frac{1 + 3(L_w \omega)^2}{[1 + (L_w \omega)^2]^2}
\end{equation}

In these equations, \(\Phi_u(\omega)\), \(\Phi_v(\omega)\), and \(\Phi_w(\omega)\) are the power spectral densities for the longitudinal, lateral, and vertical turbulence components, respectively, where $\omega$ is the frequency; \(\sigma_u\), \(\sigma_v\), and \(\sigma_w\) are the root mean square (RMS) turbulence velocities, and \(L_u\), \(L_v\), and \(L_w\) are the scale lengths for these components.

Aircraft control inputs are defined as:

\begin{equation}
a = [c_e, c_a, c_r, c_T] \in [-1, 1]^3 \times [0, 1],
\end{equation}
where \( c_a \) is the aileron deflection, \( c_e \) is the elevator deflection, \( c_r \) is the rudder deflection, and \( c_T \) is the throttle position. The deflection of the control surfaces is denormalized depending on the aircraft model used for simulation and its real characteristics.

The UAV model used in this work to obtain the results is kept confidential, but all calculations can be performed using any other JSBSim aircraft models or self-developed models. Altitude, heading, and airspeed values are normalized by a given factor for each one.

The UAV model includes not only the aerodynamics and propulsion models but also an internal flight control system (FCS) acting as a low-level Fly-By-Wire (FBW) system. This allows the user to control the angular rates and load factor, as in a real manned aircraft, instead of directly controlling the deflection of the control surfaces. Thus, the FBW system maps the pilot control inputs defined as $[\delta_e, \delta_a, \delta_r, \delta_T]$ to the control surfaces and throttle $[c_e, c_a, c_r, c_T]$ inputs, allowing the agent to control the UAV as a human pilot would.

\subsection{Reinforcement Learning Algorithms Overview}

Reinforcement Learning (RL) is one of the three primary Machine Learning (ML) paradigms, focused on maximizing cumulative reward through trial and error. Although there are various approaches within RL, they can be broadly categorized into model-free and model-based algorithms.

\subsubsection{Model-free RL}

Model-free RL is a widely used approach where the learning algorithm directly interacts with the environment to learn an optimal policy that maps observations to actions. This method does not involve learning a model of the environment's dynamics. One common family of model-free algorithms is the Actor-Critic (AC) methods, which combine the benefits of both value-based and policy-based approaches.

In Actor-Critic algorithms, the \textit{actor} updates the policy $\pi(a|s;\theta)$ based on feedback from the \textit{critic}, which evaluates the action by estimating the value function $V(s;\omega)$ or the action-value function $Q(s, a;\omega)$. The objective is to minimize the loss function for the critic, such as:

\begin{equation}
    L(\omega) = \mathbb{E}[(R_t - V(s_t;\omega))^2]
\end{equation}

and update the actor by optimizing the expected return:

\begin{equation}
  \nabla_\theta J(\theta) = \mathbb{E}[\nabla_\theta \log \pi(a|s;\theta) Q(s,a;\omega)]  
\end{equation}

Different algorithms apply diverse loss functions and update methods to improve either speed, sample efficiency, exploration capabilities, or allow continuous or discrete actions. Model-free algorithms can be further classified into on-policy and off-policy methods. On-policy methods, like PPO (Proximal Policy Optimization) \cite{ppo}, update the policy based on the actions actually taken, whereas off-policy methods, such as SAC (Soft Actor-Critic) \cite{sac} and TQC (Truncated Quantile Critics) \cite{kuznetsov2020controlling}, learn the value of the optimal policy independently of the agent's actions by using experiences from a replay buffer. The latter approach is more sample efficient, making these algorithms a better approach for continuous control tasks in some cases.

\subsubsection{Model-based RL}

Model-based RL, in contrast, involves learning a model of the environment's dynamics and exploiting it to select the best option. The model typically consists of a transition function $\hat{T}(s'|s, a)$ and a reward function $\hat{R}(s, a)$. By simulating the environment using this learned model, the agent can predict future states and rewards, allowing for more efficient learning of an actor network or planning.

A prominent example of a model-based algorithm is Dreamer, specifically the most recent version DreamerV3, developed by Danijar Hafner \cite{dreamer}. Dreamer is capable of learning long-horizon behaviors purely by latent imagination, meaning that it learns an embedded representation of the real environment and uses this embedded representation to learn the optimal policy. Moreover, it is an off-policy algorithm, so it learns from previous experiences gathered with a different policy than the one the agent is trying to learn.

\begin{figure}[ht!]
    \centering
    \includegraphics[width=\textwidth]{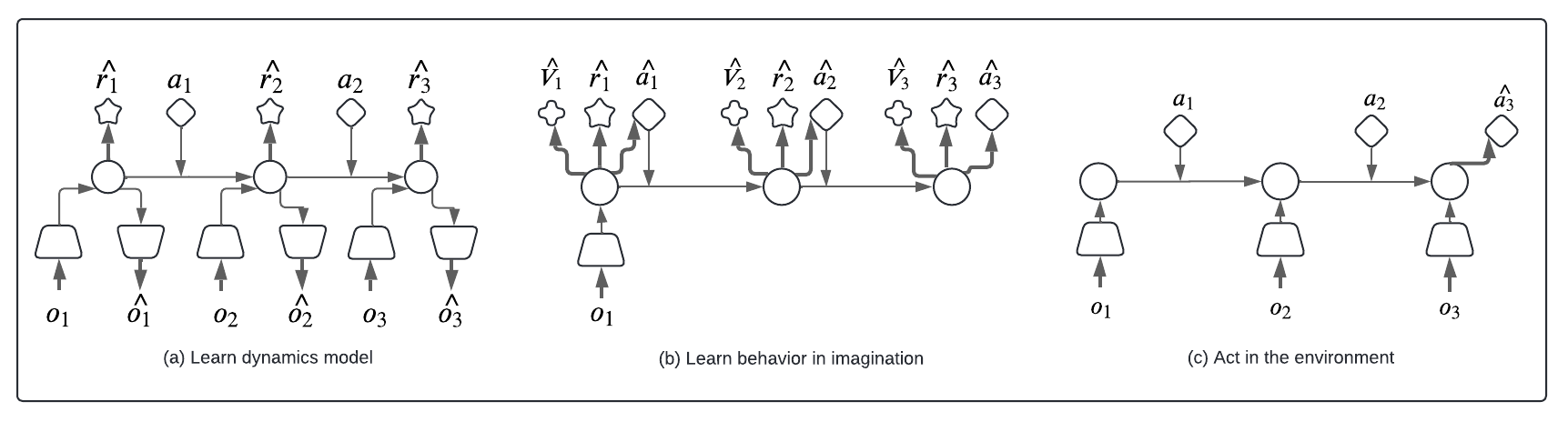}
    \caption{Diagram of the Dreamer algorithm. The agent learns a world model using observed data, then uses this model for imagination-based training of the actor and critic networks. Finally, the trained policy is applied in the real environment.}
    \label{fig:dreamer_scheme}
\end{figure}

Training Dreamer consists of three phases, each shown in Fig. \ref{fig:dreamer_scheme}. First, the world model is learned using multiple neural networks, as shown in the first part. Then, using the world model and purely imagination in the created latent representation of the observations, the actor and the critic are trained, as shown in the second part. After that, the trained policy is used to act in the real environment, with the world model still needed to embed the observations during the inference phase. In Fig. \ref{fig:dreamer_scheme}, $o$ represents the real observation from the environment, $\hat{o}$ is the reconstructed observation from the decoder, $\hat{r}$ and $\hat{V}$ are the predicted reward and value, respectively. $\hat{a}$ and $a$ are the actions taken by the agent and the actions from the buffer, respectively.

One of the most relevant aspects of Dreamer is the world model learning. The world model learns compact representations of sensory inputs through autoencoding and enables planning by predicting future representations and rewards for potential actions. The world model is implemented as a Recurrent State-Space Model (RSSM), as shown in the first diagram of Fig. \ref{fig:dreamer_scheme}. First, an encoder maps sensory inputs $x_t$ to stochastic representations $z_t$. Then, a sequence model with recurrent state $h_t$ predicts the sequence of these representations given past actions $a_{t-1}$. The concatenation of $h_t$ and $z_t$ forms the model state from which we predict rewards $r_t$ and episode continuation flags $c_t \in \{0, 1\}$ and reconstruct the inputs to ensure informative representations:

\begin{equation} \label{world_model_eq}
\begin{aligned}
&\text{Sequence model:} \quad & h_t &= f_\theta(h_{t-1}, z_{t-1}, a_{t-1}) \\
&\text{Encoder:} \quad & z_t &\sim q_\phi(z_t \mid h_t, x_t) \\
&\text{Dynamics predictor:} \quad & \hat{z}_t &\sim p_\theta(\hat{z}_t \mid h_t) \\
&\text{Reward predictor:} \quad & \hat{r}_t &\sim p_\theta(\hat{r}_t \mid h_t, z_t) \\
&\text{Continue predictor:} \quad & \hat{c}_t &\sim p_\theta(\hat{c}_t \mid h_t, z_t) \\
&\text{Decoder:} \quad & \hat{x}_t &\sim p_\theta(\hat{x}_t \mid h_t, z_t) \\
\end{aligned}
\end{equation}

The RSSM is composed of the first three expressions shown in \ref{world_model_eq}. All of these expressions form the world model of Dreamer. For a deeper explanation of the world model and actor-critic training and the architecture of Dreamer, please refer to \cite{dreamer, hafner2020mastering, hafner2019dream}.

\subsection{Markov Decision Process Formulation of the Problem} \label{mdp}

Reinforcement Learning relies on the Markov Decision Process (MDP) formulation, making it necessary to design the problem at hand as a sequence of states ($S$), actions ($A$), and rewards ($R$).

We define the problem as a fully observable MDP during training, assuming all necessary states are known and observable in the state space. This way, the agent can completely observe its own state and the necessary states of the target to determine its relative position.

\subsubsection{State Space Formulation}

The agent's state space, defined as $S$ in the MDP, comprises all necessary observations to be aware of its attitude, velocity, and position, which define the complete state of an aircraft, as well as the observations needed to know the relative position and velocity of the target, allowing it to determine the target's movement.

All the observations are assumed to be obtained from typical UAV sensors and systems and introduced to the agent as a one-dimensional vector, $o$. We assume that these observations have no noise during training and that there are no external wind perturbations, considering a calm atmosphere.

The observations corresponding to the pursuer UAV are composed of its altitude ($h$), its attitude values expressed in Euler angles ($\theta, \phi, \psi$) and angular rates ($p, q, r$), its velocity vector ($V_N, V_E, V_D$) in earth-centered coordinates, its aerodynamic values consisting of Mach number ($M_{\infty}$), angle of attack ($\alpha$), and sideslip angle ($\beta$). Additionally, load factor accelerations ($n_x, n_y, n_z$) and current angle deflections of the control surfaces and throttle ($c_e, c_a, c_r, c_T$) are included.

These proprioceptive observations give the self-observation vector:

\begin{equation}
\mathbf{o_{self}} = \begin{bmatrix}
    h \\
    \theta \\
    \phi \\
    \psi \\
    p \\
    q \\
    r \\
    V_N \\
    V_E \\
    V_D \\
    M_{\infty} \\
    \alpha \\
    \beta \\
    n_x \\
    n_y \\
    n_z \\
    c_e \\
    c_a \\
    c_r \\
    c_T
\end{bmatrix}
\end{equation}

Regarding the target-related observations, the agent can observe several relative parameters from the evader UAV. These observations are composed of the relative altitude error between the two UAVs ($\epsilon_h$), the relative angles: Aspect Angle (AA), Antenna Train Angle (ATA), and Heading Cross Angles (HCA) (how these angles are measured is shown in Fig. \ref{fig:angles_ATA_AA}); the relative velocity measured from the velocity of the pursuer and the evader ($\epsilon_{V_N}, \epsilon_{V_E}, \epsilon_{V_D}$), and the relative distance measured between the two UAVs ($d_{\text{rel}}$). This relative distance is measured using the Haversine formulas shown below:

\begin{figure}[ht!]
    \centering
    \includegraphics[width=0.4\textwidth]{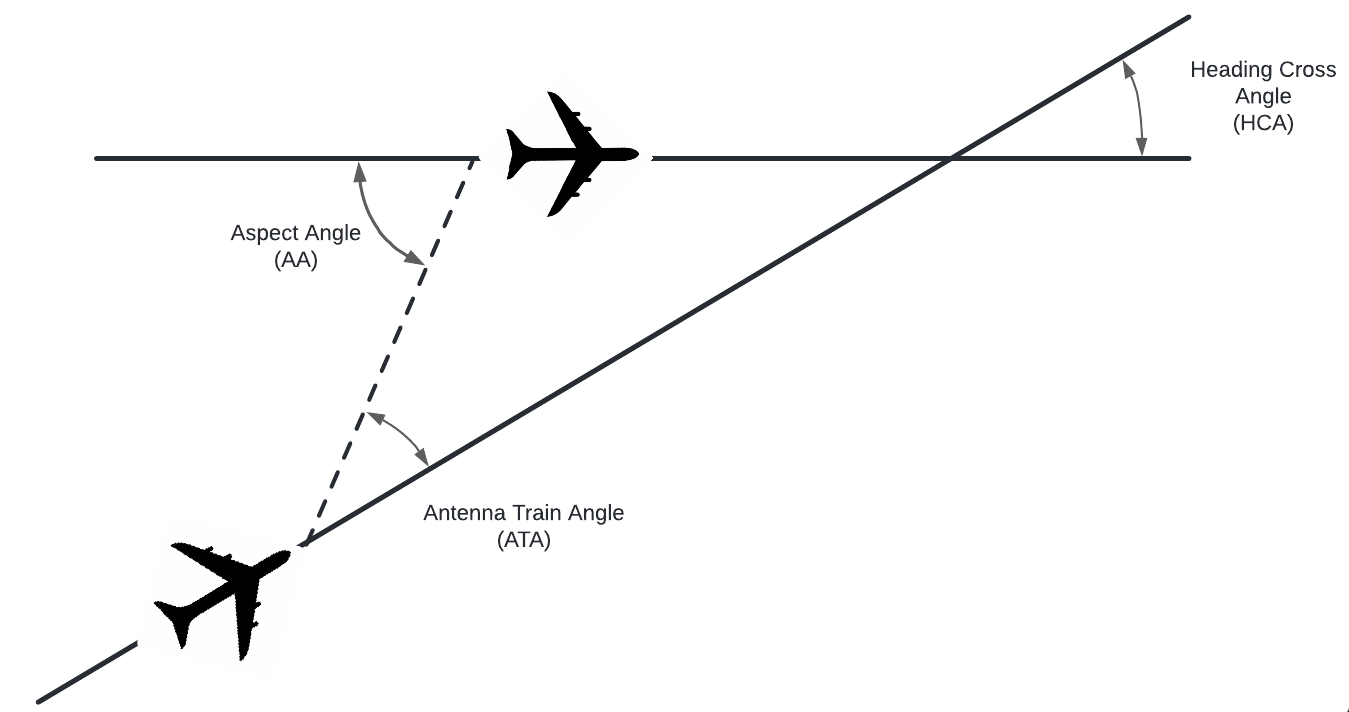}
    \caption{Diagram showing the Aspect Angle (AA) and Antenna Train Angle (ATA) between the pursuer and target UAVs.}
    \label{fig:angles_ATA_AA}
\end{figure}

\begin{equation}
a = \sin^2\left(\frac{\Delta \phi}{2}\right) + \cos(\phi_1) \cos(\phi_2) \sin^2\left(\frac{\Delta \lambda}{2}\right)
\end{equation}

\begin{equation}
c = 2 \tan^{-1}\left( \sqrt{a}, \sqrt{1-a} \right)
\end{equation}

\begin{equation}
d_{\text{rel}} = R \cdot c
\end{equation}

where:
\begin{itemize}
    \item $\phi_1$ and $\phi_2$ are the latitudes of the two UAVs in radians,
    \item $\Delta \phi$ is the difference in latitude,
    \item $\Delta \lambda$ is the difference in longitude,
    \item $R$ is the Earth's radius,
    \item $d_{\text{rel}}$ is the relative distance between the two UAVs.
\end{itemize}

These target-related observations give the target observation vector:

\begin{equation}
\mathbf{o}_{\text{relative}} = \begin{bmatrix}
    \epsilon_h \\
    \text{AA} \\
    \text{ATA} \\
    \text{HCA} \\
    \epsilon_{V_N} \\
    \epsilon_{V_E} \\
    \epsilon_{V_D} \\
    d_{\text{rel}}
\end{bmatrix}
\end{equation}

The concatenation of these two vectors forms the state space, $S$, for the agent, providing the observation, $o$, at each time step. Before being fed to the agent, these observations undergo a normalization transformation. The normalization process is described as follows:

\begin{itemize}
    \item Angle measurements are transformed to their sine and cosine values. This transformation ensures the values are normalized between -1 and 1, preserving all information.
    \item Altitude, velocity vector components, Mach number, and angular rates are normalized using the min-max method.
    \item Load factors, control surface deflections, and throttle are inherently normalized by default in the JSBSim Flight Dynamics Model (FDM) between -1 and 1.
    \item Relative distance ($d_{\text{rel}}$) is normalized using the formula $d_{\text{norm}} = \frac{1}{1 + d_{\text{rel}}}$.
\end{itemize}

Table \ref{table:normalization} summarizes the normalization parameters.

\begin{table*}[ht!]
\centering
\begin{tabular}{cccc}
\hline
\textbf{Parameter} & \textbf{Normalization Method} \\
\hline
Angle Measurements & Sine and Cosine \\
\hline
Altitude & Min-Max \\
\hline
Velocity Components & Min-Max\\
\hline
Mach Number & Min-Max\\
\hline
Angular Rates & Min-Max\\
\hline
Load Factors & - \\
\hline
Control Surface Deflections & - \\
\hline
Throttle &  - \\
\hline
Relative Distance ($d_{\text{rel}}$) & $d_{\text{norm}} = \frac{1}{1 + d_{\text{rel}}}$ \\
\hline
\end{tabular}
\caption{Normalization Parameters for Observations}
\label{table:normalization}
\end{table*}

\subsubsection{Action Space Formulation} \label{section:action_space}

As mentioned in Section \ref{section:simulation_env}, the UAV model used in JSBSim for the simulations includes a Fly-By-Wire system model which maps the pilot's commanded inputs to the control surface deflections and throttle. Thus, the RL agent needs to take the same actions as a human pilot would. The action space is defined as follows:

\begin{equation}
A = [\delta_e, \delta_a, \delta_r, \delta_T] \in [-1, 1]^3 \times [0, 1],
\end{equation}

These actions are defined as:

\begin{itemize}
    \item $\delta_e$: pitch rate command to control the longitudinal motion of the UAV.
    \item $\delta_a$: roll rate command to control the coupled lateral/directional motion of the UAV.
    \item $\delta_r$: yaw rate command to control the coupled lateral/directional motion of the UAV.
    \item $\delta_T$: throttle control, which is direct.
\end{itemize}

All control commands in the action space ($A$) are normalized between -1 and 1 by default in the FDM. Actions taken by the agent at each time step are introduced directly to the Flight Dynamics Model simulation.

\subsubsection{Reward Function Definition}

A goal-oriented reward function is defined for this problem, where the goal is the event when the dynamic target is reached by the agent, i.e., the agent is within a goal distance ($d_{goal}$) of the evader.

\begin{figure}[ht!]
    \centering
    \includegraphics[width=0.45\textwidth]{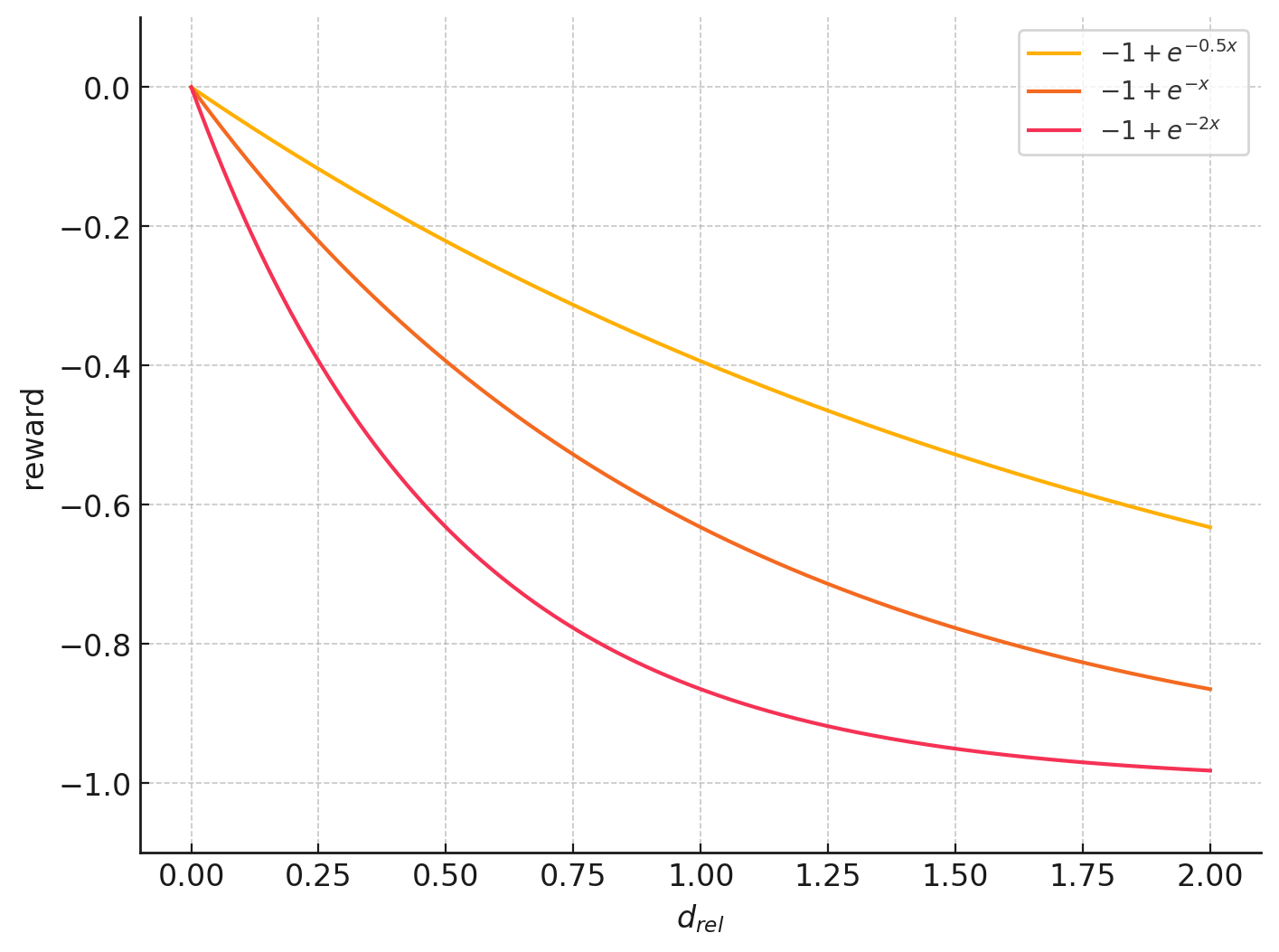}
    \caption{Reward shaping function for different values of the shaping parameter $k$.}
    \label{fig:reward_function}
\end{figure}

Given the problems of sample inefficiency derived from the goal-oriented reward—since the agent only receives a reward when the objective is reached—a shaping reward/penalization is added to help the agent understand the task objective in fewer time steps. This shaping penalization function can be observed in Fig. \ref{fig:reward_function} for different shaping parameters $k$. We found that using a parameter $k=1$ improves the sample efficiency of agents during training for this task.

Additionally, a penalization is added when the pursuer UAV is below 2000 ft of altitude to avoid simulation errors in the FDM due to ground contact. This condition is also an episode termination condition, which will be detailed in the next section.

Finally, the piecewise reward function for the pursuer agent is defined as shown in Equation \ref{equ:reward_def}.

\begin{equation} \label{equ:reward_def}
f(d_{\text{rel}}, h) = 
\begin{cases} 
-1 + e^{-k \cdot d_{\text{rel}}} + 2000 & \text{if } d_{\text{rel}} < d_{\text{goal}} \\
-1 + e^{-k \cdot d_{\text{rel}}} - 2000 & \text{if } h < 2000 \, \text{ft} \\
-1 + e^{-k \cdot d_{\text{rel}}} & \text{otherwise}
\end{cases}
\end{equation}

\subsubsection{Training Framework}

Training is designed based on the Gymnasium framework architecture. The framework is divided into the simulation part and the task-related part. The simulation part is carried out by the Flight Dynamics Model, handling the transfer of information with the task-related part to provide the dynamic values calculated in the simulator. The task-related part handles all the training setup, episode initialization, evasion strategies of the evader, and the calls to the running FDM simulation of the evader, which is independent of the pursuer agent.

At the beginning of each episode, the training is set up as follows:

\begin{itemize}
    \item The target UAV is placed in a randomly sampled position given by its latitude ($\phi$), longitude ($\lambda$), and altitude ($h$).
    \item The agent UAV, the pursuer, is placed randomly but depends on the previously placed evader's position. Table \ref{table:pursuer_initialization} shows the range of values added to the target's position to calculate the position of the pursuer. The latitude and longitude of the pursuer based on the distance and bearing to the evader are calculated using Equation \ref{equ:position_calculation}.
    \item At the start of the episode, an evasion strategy is randomly selected for the evader.
\end{itemize}

\begin{equation} \label{equ:position_calculation}
\begin{aligned}
    \phi_2 &= \arcsin(\sin(\phi_1) \cdot \cos\left(\frac{d}{R}\right) \notag \\
    &\quad + \cos(\phi_1) \cdot \sin\left(\frac{d}{R}\right) \cdot \cos(\theta)) \notag \\
    \lambda_2 &= \lambda_1 + \arctan2\left(\sin(\theta) \cdot \sin\left(\frac{d}{R}\right) \cdot \cos(\phi_1)\right.,  \\
    &\quad \left. \cos\left(\frac{d}{R}\right) - \sin(\phi_1) \cdot \sin(\phi_2)\right)
\end{aligned}
\end{equation}

\noindent Where:
\begin{itemize}
    \item $\phi_1$: Target's latitude in radians.
    \item $\lambda_1$: Target's longitude in radians.
    \item $\theta$: Relative bearing in radians (direction to move in, measured clockwise from north).
    \item $d$: Distance between the two UAVs.
    \item $R$: Radius of the Earth.
\end{itemize}

\begin{table}[ht!]
\centering
\begin{tabular}{cc}
\hline
Altitude ($h$)   & -3 to 3 \\ \hline
Distance ($d$)  & 1 to 6       \\ \hline
Bearing ($\Psi$) & 0 to 36       \\ \hline
\end{tabular}
\caption{Normalized values for pursuer position placement based on the position of the evader.}
\label{table:pursuer_initialization}
\end{table}

\begin{figure}[h]
    \centering
    \includegraphics[width=0.48\textwidth]{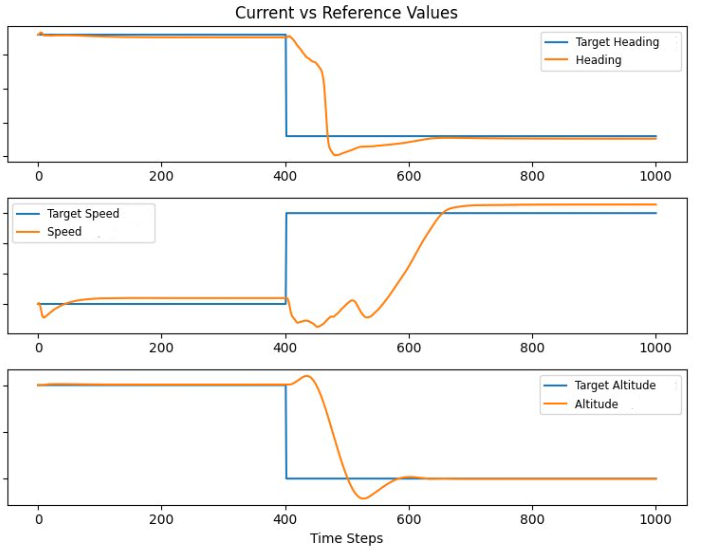}
    \caption{Example of setpoint controller policy usage for the evader UAV tracking reference values for altitude, heading, and airspeed.}
    \label{fig:setpoint_controller}
\end{figure}

To control the evader UAV, a policy trained through Reinforcement Learning is used as a controller. This policy maps from the concatenation of the aircraft's internal state and the goal state to the pilot-related commands as mentioned in Section \ref{section:action_space} for the pursuer agent. This goal state is given by the commanded setpoint values in altitude ($h_{ref}$), heading ($\Psi_{ref}$), and airspeed ($V_{T_{ref}}$), meaning that the objective of the controller policy is to minimize the deviations between the current values and the setpoint values. Fig. \ref{fig:setpoint_controller} showcases an example of how this evader control policy tracks the reference values for $h_{ref}$, $\Psi_{ref}$, and $V_{T_{ref}}$ simultaneously.

Table \ref{table:evader_control_range} shows the envelope of values where the controller policy has been trained and can be used during inference. This range of values is used to command the evader's setpoints in the evasion strategies.

\begin{table}[ht!]
\centering
\begin{tabular}{cc}
\hline
Altitude ($h$)   & 3 to 15 \\ \hline
Heading ($\Psi$) & 0 to 36        \\ \hline
Airspeed ($V_T$)   & 3 to 12       \\ \hline
\end{tabular}
\caption{Range of normalized values for controlling the setpoint policy of the evader.}
\label{table:evader_control_range}
\end{table}

Evasion strategies are defined by controlling the dynamics of the evader through commands in these three reference values, $h_{ref}$, $\Psi_{ref}$, and $V_{T_{ref}}$, partially based on the dynamics of the pursuer, even though both UAV dynamics run on different simulations.

During training, the commanded setpoints which define the evasion maneuvers are selected based on two strategies.

\textbf{\textit{First evasion strategy:}} This evasion strategy is based on time, changing the commanded setpoints to the evader at a fixed number of time steps. For the particular case of the training in this work, the commanded references to the evader change every 50 time steps. Commanded heading ($\Psi_{ref}$) and altitude ($h_{ref}$) are chosen based on the corresponding values of the pursuer at that time step, adding a given quantity randomly sampled from a range of possible values. The commanded airspeed ($V_{T_{ref}}$) is independent of the pursuer's airspeed and is also randomly sampled from a range.

\textbf{\textit{Second evasion strategy:}} Unlike the first strategy, this one is based on the distance to the pursuer, producing a change in the commanded values when the pursuer is closer than a limit distance to the evader. In this work, this distance is chosen to be 2 km. This way, the evader UAV's dynamics are changed at every time step while the pursuer is in the distance range. Commanded values are selected in the same way as in the first strategy. Table \ref{table:range_values_strategies} shows the range of values that can be added to the pursuer's dynamics to define the commanded values for both strategies.

\begin{table}[ht!]
\centering
\begin{tabular}{cc}
\hline
Altitude ($h$)   & -2 to 2 \\ \hline
Heading ($\Psi$) & -4 to 4       \\ \hline
Airspeed ($V_T$)   & 7 to 12        \\ \hline
\end{tabular}
\caption{Range of normalized values for evasion strategies definition based on the dynamics of the pursuer.}
\label{table:range_values_strategies}
\end{table}

Training is set up with the following configuration:

\begin{itemize}
    \item The maximum length of episodes is 2000 steps.
    \item The simulation integration frequency is 60 Hz.
    \item The control frequency is 12 Hz (5 simulator steps per agent action).
    \item The maximum training budget is 1M steps for each algorithm.
\end{itemize}

Episodes termination conditions are defined as:

\begin{itemize}
    \item Episode is truncated at the maximum length of 2000 steps, without penalization.
    \item Episode is terminated with penalization if the agent is lower than 2000 ft, to avoid ground contact errors in the simulation.
    \item Episode is terminated with reward if the target is caught.
\end{itemize}

\subsubsection{Validation Scenarios} \label{sec:validation_scenarios}

To assess the capabilities of the trained agents to act against unseen evader behaviors during training, two different evasion strategies are included during validation:

\begin{itemize}
    \item \textbf{\textit{Random evasion:}} A randomly selected setpoints strategy is defined to test the agents against unpredictable and irrational target movements.
    \item \textbf{\textit{User-controlled evasion:}} During validation, the possibility for a user to control the commanded values in the evader through $h_{ref}$, $\Psi_{ref}$, and $V_{T_{ref}}$ is defined to test the generalization capabilities of the trained agents to new evasion maneuvers.
\end{itemize}

Additionally, two different test scenarios are defined to evaluate the robustness of the trained agents:

\begin{itemize}
    \item \textbf{\textit{Wind gust perturbations:}} Wind external perturbations are added during validation to test robustness, using the Dryden Gusts model defined in Section \ref{section:simulation_env}. The wind gust magnitudes introduced to the simulation can be seen in Table \ref{tab:wind_gust_severe}. These wind gusts directly affect the dynamics of the UAV calculated in the FDM, and the agent is unaware of these changes in the dynamics in a direct way.
    \item \textbf{\textit{Sensor noise perturbations:}} Sensor noise is added to the observations seen by the agent during validation to assess robustness against instabilities in the observations. The noise added for validation follows a Gaussian distribution with a mean of zero and standard deviation of 0.25 ($N(0, 0.25)$), considering that the observations are already normalized.
\end{itemize}

\begin{table}[ht!]
\centering
\begin{tabular}{cc}
\hline
Component & Distribution \\ \hline
$W_N$ (North) & $N(0, 45)$ \\ \hline
$W_E$ (East) & $N(0, 45)$ \\ \hline
$W_D$ (Down) & $N(0, 45)$ \\ \hline
\end{tabular}
\caption{Gaussian distribution parameters for the components of the Wind Gust vector in a severe situation, represented in ft/s and composing the vector ($W_N$, $W_E$, $W_D$). These parameters are used to randomly generate the gust values for each time step.}
\label{tab:wind_gust_severe}
\end{table}

\section{Results} \label{results}

In this section, we present the results obtained during this work, separating between the training results and the assessment of the trained algorithms during validation in different scenarios presented in Section \ref{sec:validation_scenarios}.

\subsection{Training Results}

Our goal was to train and compare model-based and model-free RL algorithms, both in terms of performance and sample efficiency during training. To assess the latter, we set a predefined training budget both for hyperparameter tuning and for final training. These training budgets are 10M steps for hyperparameter tuning and 1M time steps for final training. All algorithms were trained using the same set of random seeds.

We consider an algorithm to have accomplished the task if, within the given budget, the agent is able to learn the final goal of catching the moving target before the episode ends. Given that the only positive reward is received upon catching the target, the theoretical maximum episode return is 2000, indicating that any episode return greater than zero means the target was caught.

Several algorithms were tried during this work, including well-known RL algorithms such as PPO (Proximal Policy Optimization) \cite{ppo}, DDPG (Deep Deterministic Policy Gradient) \cite{lillicrap2015continuous}, and TD3 (Twin Delayed DDPG) \cite{fujimoto2018addressing}. However, these three algorithms were unable to learn the task within the given budget (i.e., they did not surpass zero episode return), failing to find the correct set of hyperparameters to accomplish the task. This behavior is expected given that the task at hand is complex and goal-oriented, producing sparse rewards during training despite a shaping reward being added to help with sample efficiency. Considering that PPO is an on-policy method (and overall less sample efficient), and DDPG and TD3 are very sensitive to hyperparameters (a common problem in RL), it is expected that these algorithms could not be trained successfully under the given conditions and task.

Nevertheless, we found three algorithms that were able to accomplish the task under the given conditions. On the model-based side, DreamerV3 was chosen due to its excellent performance across a wide set of tasks, being state-of-the-art in some \cite{dreamer}. On the model-free side, SAC (Soft Actor-Critic) \cite{sac} and TQC (Truncated Quantile Critics) \cite{kuznetsov2020controlling}, which is inspired by SAC, were able to learn the task within the given budget.

Hyperparameter tuning was applied to both model-free algorithms, TQC and SAC. After tuning, the SAC algorithm was trained using both its own found hyperparameters and those from TQC, with better results observed using the latter. These hyperparameters are shown in Table \ref{table:tqc_hyperparameters}. The model-based algorithm, DreamerV3, did not require hyperparameter tuning (as stated in the original paper), so default hyperparameters were used. For these hyperparameters, please refer to \cite{dreamer}.

It should be noted that the SAC algorithm was not trained under the same conditions as DreamerV3 and TQC. Since SAC was unable to learn the task starting from the established training setup, a manual curriculum was added to SAC training, pre-training the algorithm for 500k time steps on a fixed line trajectory target strategy to facilitate learning. After that pre-training, regular training was applied as for the other two algorithms. This implies SAC was trained for a total of 1.5M time steps.

\begin{table}[h]
\centering
\begin{tabular}{cc}
\hline
\textbf{Hyperparameter} & \textbf{Value} \\ \hline
Learning rate ($\alpha$) & 3.0e-4 \\ \hline
Buffer size & 50000 \\ \hline
Batch size & 256 \\ \hline
Discount factor ($\gamma$) & 0.99 \\ \hline
Polyak averaging coefficient ($\tau$) & 0.08 \\ \hline
Training frequency & 128 \\ \hline
Gradient steps & 128 \\ \hline
Learning starts & 15000 \\ \hline
\multicolumn{2}{c}{\textbf{Policy Arguments}} \\ \hline
Log standard deviation initialization & 0.165 \\ \hline
Network architecture  & [256, 256] \\ \hline
Number of critics  & 2 \\ \hline
Activation function & "Tanh" \\ \hline
\end{tabular}
\caption{Hyperparameters for TQC and SAC algorithms.}
\label{table:tqc_hyperparameters}
\end{table}

\begin{figure*}[h]
    \centering
    \begin{subfigure}[b]{\textwidth}
        \centering
        \includegraphics[width=0.9\textwidth]{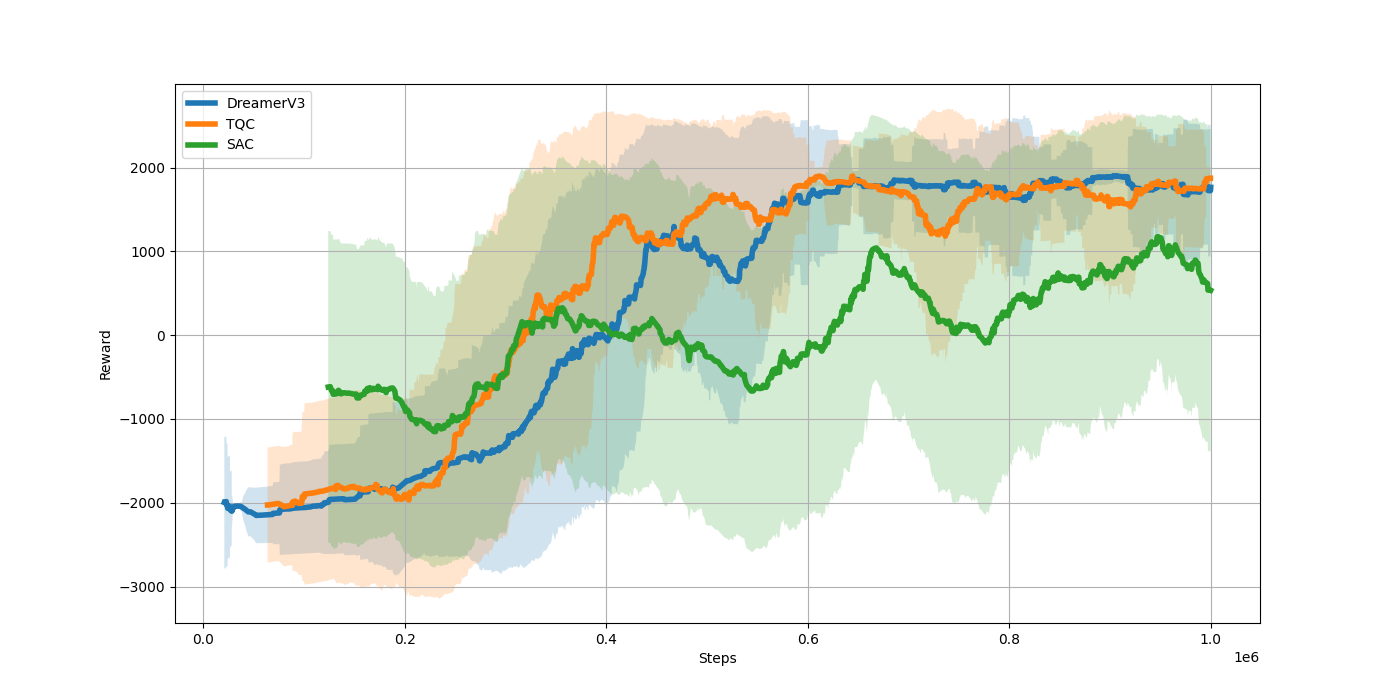}
        \caption{Episode returns per training steps.}
        \label{fig:training_rewards}
    \end{subfigure}
    \hfill
    \begin{subfigure}[b]{\textwidth}
        \centering
        \includegraphics[width=0.9\textwidth]{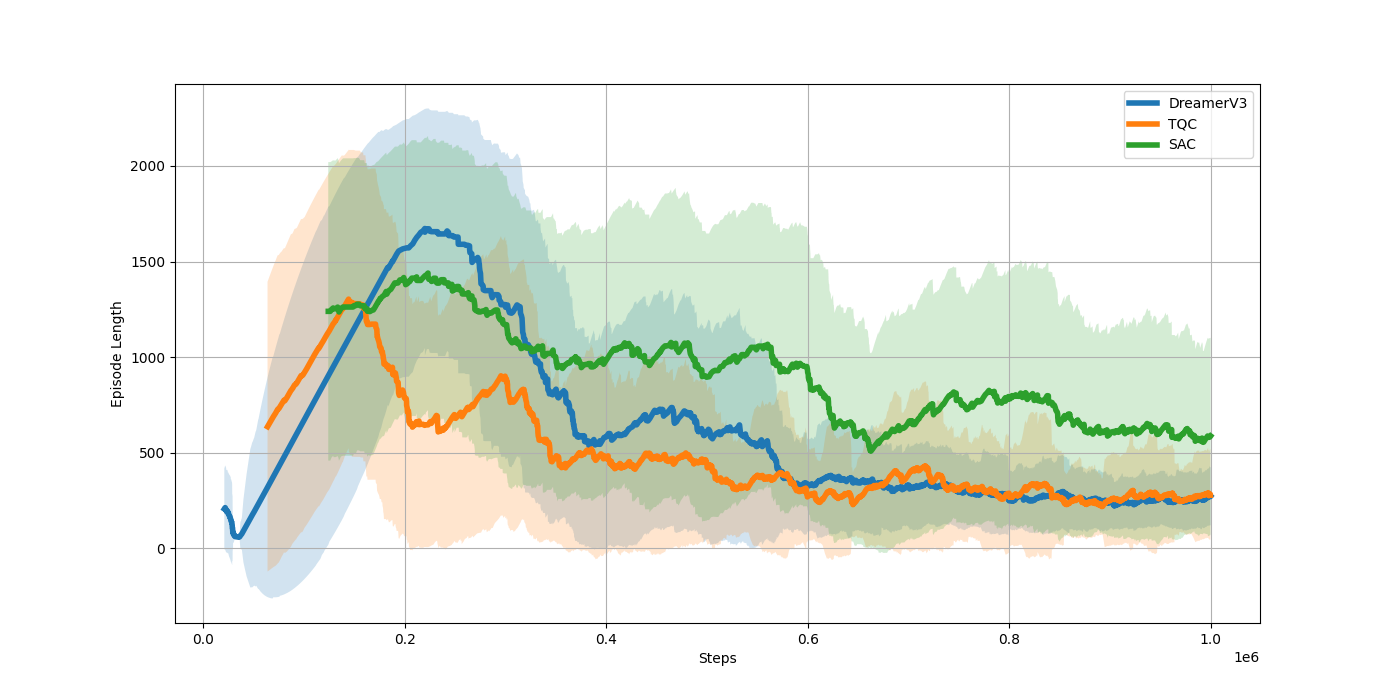}
        \caption{Episode lengths per training steps.}
        \label{fig:training_length}
    \end{subfigure}
    \caption{Comparison of returns and episode lengths per step for the Dreamer, TQC, and SAC algorithms during training. Results are presented as moving averages and moving standard deviations, each calculated with a moving window of 100 steps. The theoretical maximum return is 2000.}
    \label{fig:training_length_and_reward}
\end{figure*}

Figure \ref{fig:training_length_and_reward} shows the results for the three trained algorithms in terms of episode returns and episode length, with the goal being to maximize the first and minimize the second. A moving mean and a moving standard deviation have been applied to the results for clarity. Figure \ref{fig:training_rewards} shows the results in episode returns, indicating that DreamerV3 and TQC are able to learn the task efficiently, taking only around 300k-400k steps to surpass the zero return value (indicating that the target was intercepted). Contrary to common assumptions, in this case, the model-free algorithm, TQC, was more sample efficient than the model-based DreamerV3. However, DreamerV3 proved to be more stable during training, as evidenced by the lower standard deviation values between episodes. Both algorithms approximate the theoretical maximum reward of 2000 after 600k time steps, indicating that they are able to catch the target as quickly as possible. On the other hand, SAC, despite benefiting from pre-training, is not as stable as the other two algorithms when trying to catch the dynamic and moving target, and although it understood the task, it is not able to catch it efficiently.

Figure \ref{fig:training_length} shows the episode lengths for each algorithm. This figure illustrates the learning process for each agent. Initially, training episodes end quickly due to termination conditions, but as training progresses, agents learn to follow the target, and subsequently, to catch it as quickly as possible.

\subsection{Validation Results}

As mentioned in Section \ref{sec:validation_scenarios}, the algorithms have been tested not only against the same evasion strategies used for the evader during training but also against two different strategies: completely random and human-controlled. A validation across all these circumstances was studied for the three algorithms, DreamerV3, TQC, and SAC.

This study consists of evaluating the performance of each algorithm for each evasion strategy separately, measuring the performance in mean episode return and standard deviation across all of them. The same random seeds were used for the three algorithms to handle randomness in the environment. Evasion strategies were tested across 100 episodes.

\begin{figure*}[h]
    \centering
    \includegraphics[width=0.9\textwidth]{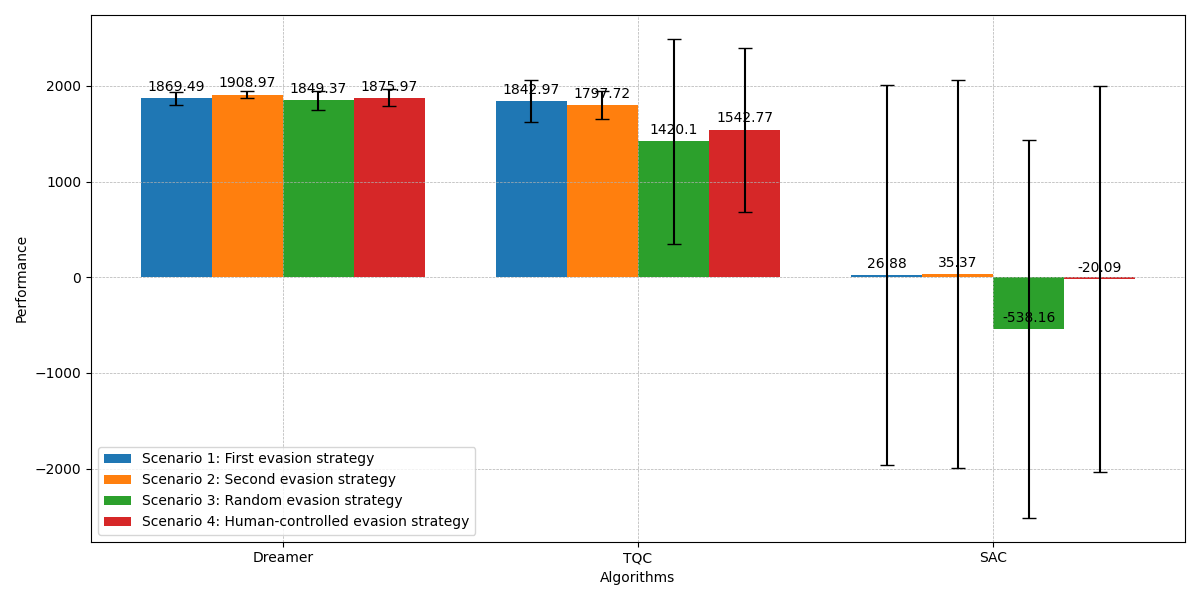}
    \caption{Validation of algorithms across different strategies.}
    \label{fig:validation_normal}
\end{figure*}

Fig. \ref{fig:validation_normal} shows the results of the validation for each algorithm across all evasion strategies, showing performance both in mean and standard deviation. As can be seen in the results, Dreamer and TQC achieve similar performance in scenarios 1 and 2, which are the training evasion scenarios, with Dreamer showcasing more stability in the decision-making process given the lower standard deviation. Both algorithms obtain excellent results in these scenarios, considering that the maximum possible reward is 2000, as mentioned before.

However, when the algorithms are tested using the two validation evasion strategies, scenarios 3 and 4, it can be seen how TQC reduces its performance and increases its standard deviation across episodes considerably, whereas Dreamer is consistent with its performance, achieving similar results as for the training strategies. This showcases Dreamer's ability to generalize well to different target behaviors beyond those seen during training. We assume that the imagination-based training framework and latent state-space of Dreamer enable it to form a more robust representation and prediction of the target.

The SAC algorithm, on the other hand, shows far worse results than Dreamer and TQC, being unable to catch the evader with a random strategy and only able to reach the target inefficiently for the other scenarios.

\begin{figure*}[h]
    \centering
    \begin{subfigure}[b]{0.45\textwidth}
        \centering
        \includegraphics[width=\textwidth]{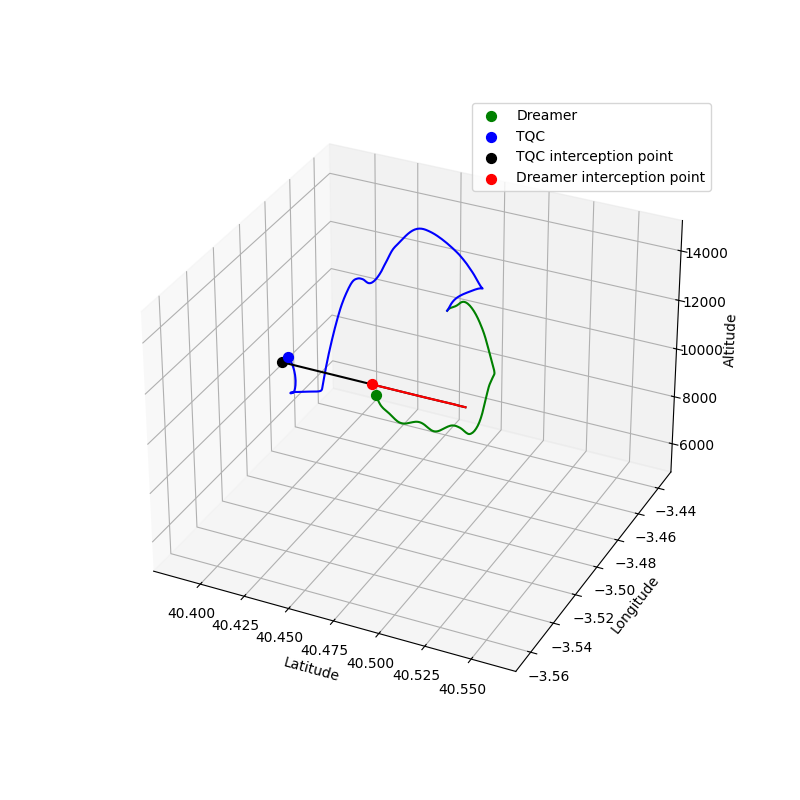}
        \caption{Trajectory comparison until intersection between Dreamer and TQC.}
        \label{fig:trajectory_comp_line_3d}
    \end{subfigure}
    \begin{subfigure}[b]{0.45\textwidth}
        \centering
        \includegraphics[width=\textwidth]{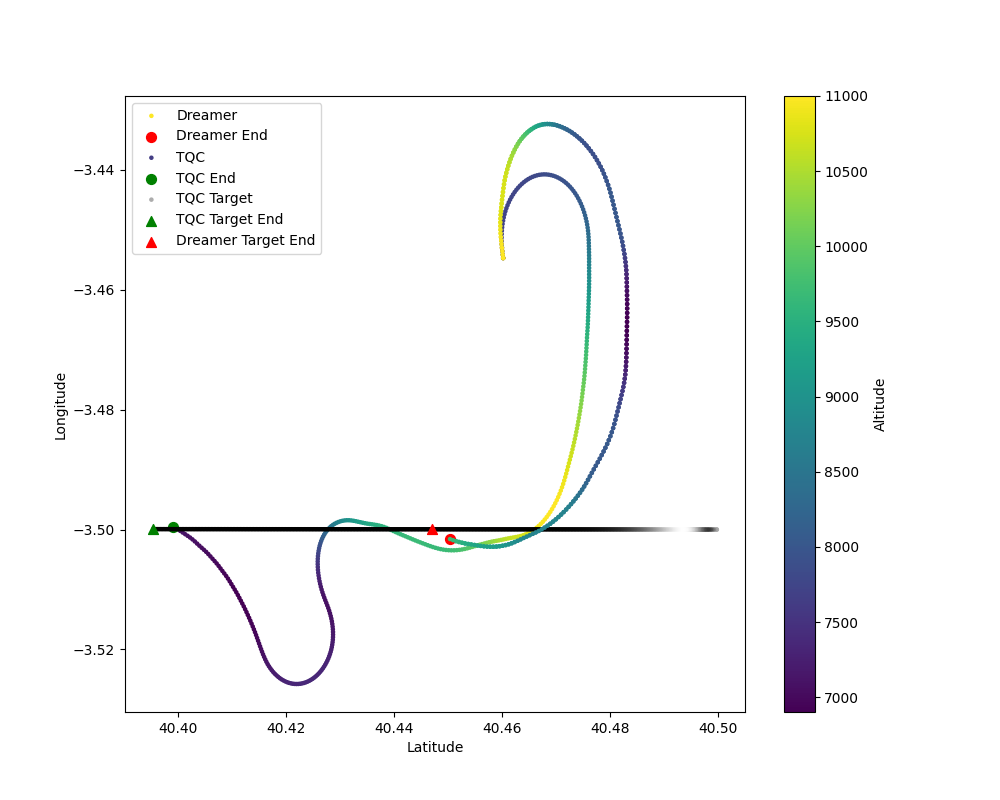}
        \caption{2D trajectory comparison until intersection between Dreamer and TQC.}
        \label{fig:trajectory_comp_line_2d}
    \end{subfigure}
    \begin{subfigure}[b]{0.45\textwidth}
        \centering
        \includegraphics[width=\textwidth]{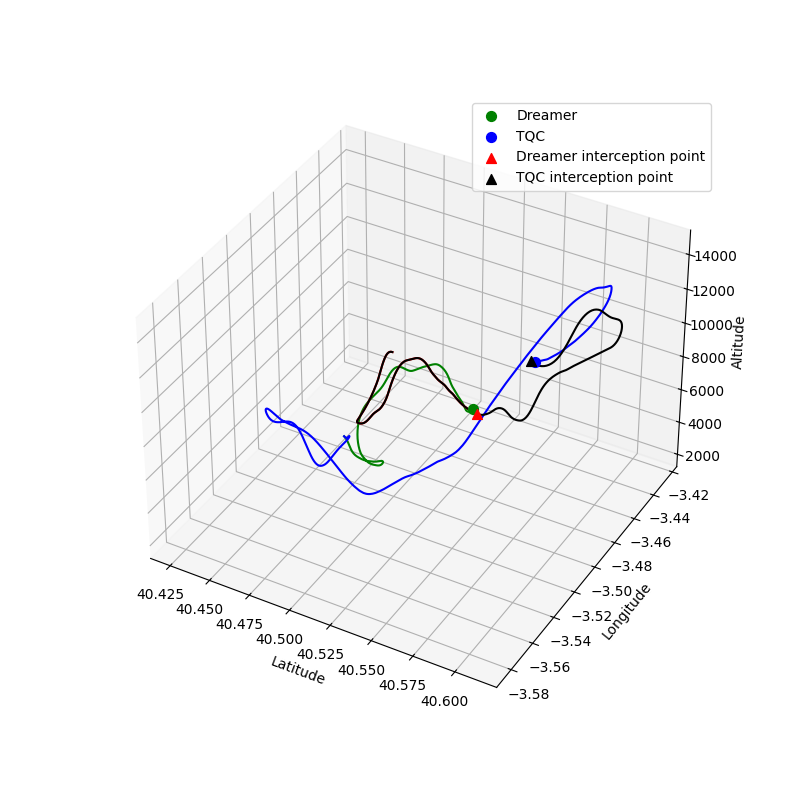}
        \caption{3D trajectory comparison until intersection between Dreamer and TQC (random evasion).}
        \label{fig:trajectory_comp_random_3d}
    \end{subfigure}
    \begin{subfigure}[b]{0.45\textwidth}
        \centering
        \includegraphics[width=\textwidth]{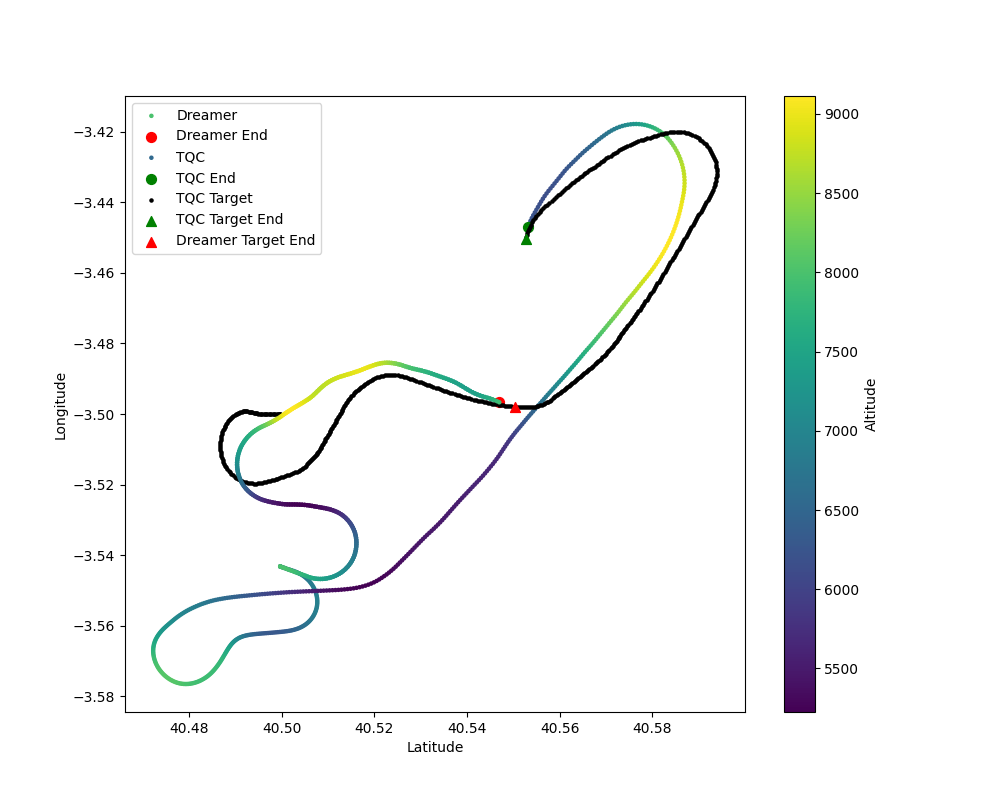}
        \caption{2D trajectory comparison until intersection between Dreamer and TQC (random evasion).}
        \label{fig:trajectory_comp_random_2d}
    \end{subfigure}
    \caption{Trajectory comparison until intersection between Dreamer and TQC.}
    \label{fig:trajectory_comp_line}
\end{figure*}

In Fig. \ref{fig:trajectory_comp_line}, an example of an episode trajectory can be seen, comparing the trajectory of the Dreamer agent with the TQC agent. In this case, the evader has no evasion strategy, resulting in a fixed line trajectory, which was not seen during training either. This example shows the same episode for both agents, starting from the same initial conditions. Fig. \ref{fig:trajectory_comp_line_3d} shows the trajectory of both agents and the target in three dimensions, represented by latitude, longitude, and altitude. Dreamer's trajectory is shown in green, with the interception point marked in red. For TQC, the agent's trajectory is shown in blue and the target's in black. It can be seen that Dreamer catches the target faster despite departing from the same point, showcasing a more efficient sequence of decisions in controlling the aircraft to reach the goal. Fig. \ref{fig:trajectory_comp_line_2d} shows the same trajectories in two dimensions, with altitude represented as a color map. Here, it is clear that the trajectory generated by the Dreamer agent's control is more optimal than that of the TQC agent.

In Figs. \ref{fig:trajectory_comp_random_3d} and \ref{fig:trajectory_comp_random_2d}, another episode example is displayed in a similar manner, but in this case, the evader follows a random evasion strategy (scenario 3). Initial conditions and the evader's actions and trajectory are the same for both algorithms (since in this case, the evader's decisions do not rely on the pursuer's position). The red point marks the intersection point of Dreamer with the target, and the black point marks that of TQC. In this case, Dreamer catches the target in 172 steps, whereas the TQC agent takes 328 steps, showing that Dreamer is again more efficient for this particular case.

\begin{figure*}[h]
    \centering
    \begin{subfigure}[b]{0.38\textwidth}
        \centering
        \includegraphics[width=\textwidth]{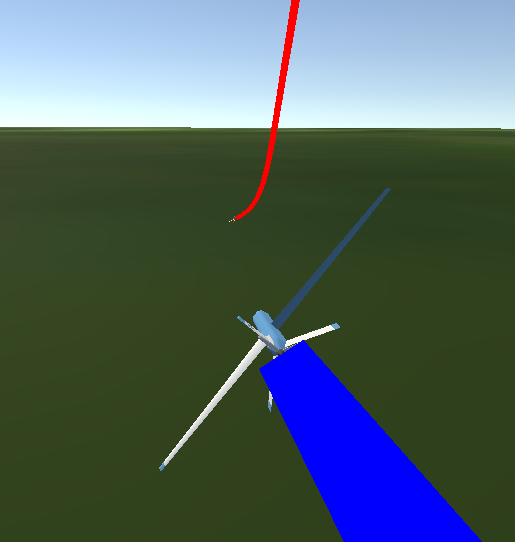}
        \caption{Pursuer's point of view.}
        \label{fig:unity_1_1}
    \end{subfigure}
    \begin{subfigure}[b]{0.38\textwidth}
        \centering
        \includegraphics[width=\textwidth]{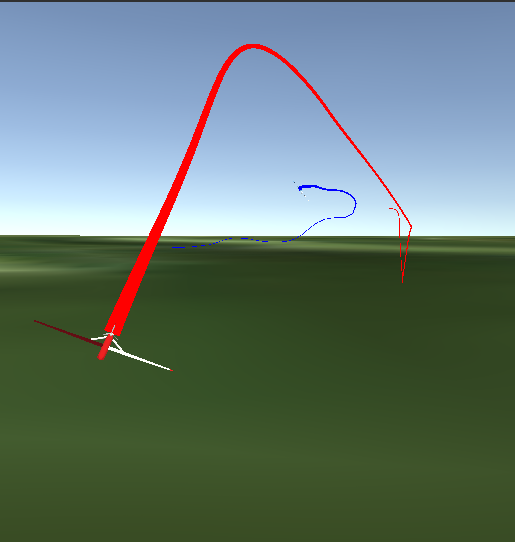}
        \caption{Trajectory of the evader and pursuer. View 1.}
        \label{fig:unity_1_2}
    \end{subfigure}
    \begin{subfigure}[b]{0.38\textwidth}
        \centering
        \includegraphics[width=\textwidth]{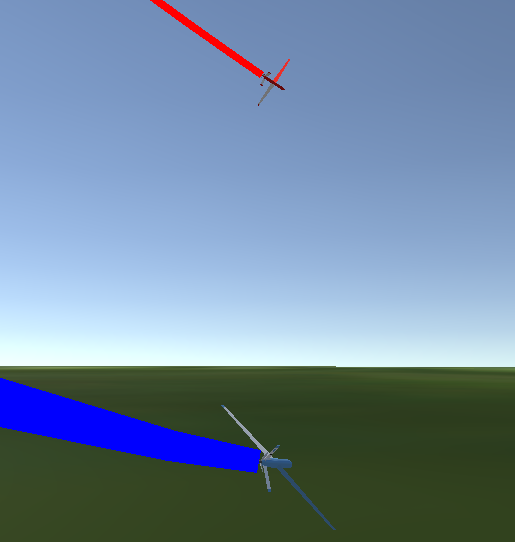}
        \caption{Trajectory of the evader and pursuer. View 2.}
        \label{fig:unity_1_3}
    \end{subfigure}
    \begin{subfigure}[b]{0.38\textwidth}
        \centering
        \includegraphics[width=\textwidth]{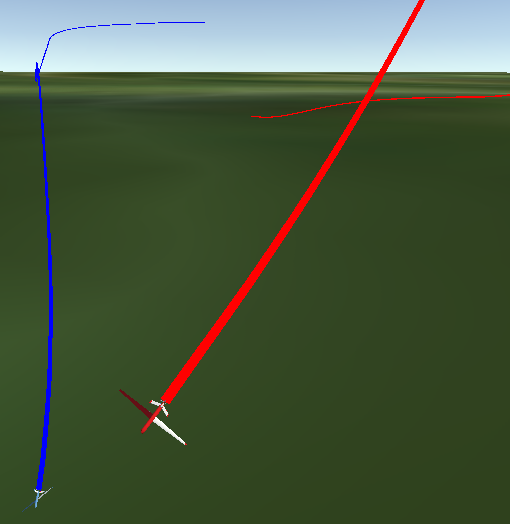}
        \caption{Trajectory of the evader and pursuer. View 3.}
        \label{fig:unity_1_4}
    \end{subfigure}
    \begin{subfigure}[b]{0.38\textwidth}
        \centering
        \includegraphics[width=\textwidth]{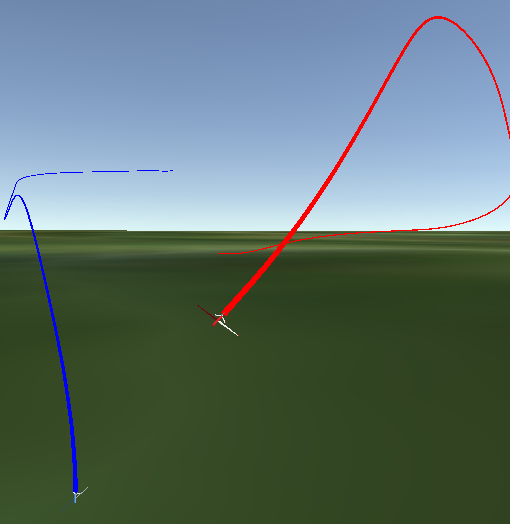}
        \caption{Trajectory of the evader and pursuer. View 4.}
        \label{fig:unity_1_5}
    \end{subfigure}
    \begin{subfigure}[b]{0.38\textwidth}
        \centering
        \includegraphics[width=\textwidth]{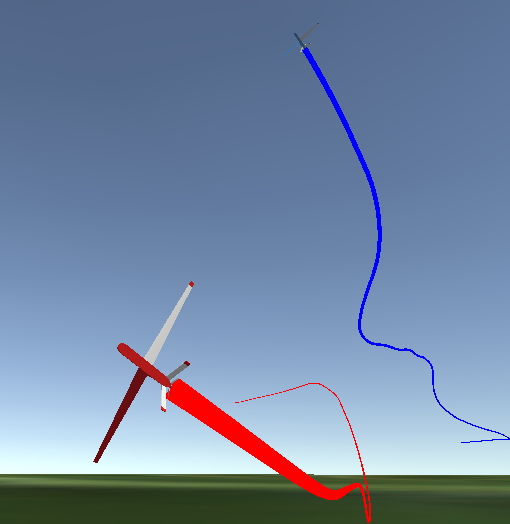}
        \caption{Trajectory of the evader and pursuer. View 5.}
        \label{fig:unity_1_6}
    \end{subfigure}

\caption{Visualization in Unity showing the trajectories of both the agent and the target across different episodes. The red UAV represents the evader (target), while the blue UAV represents the pursuer (agent). \textbf{{\textit{\href{https://youtu.be/dJd7lmK3QuU}{Link to video}}}}}
    \label{fig:unity_1}
\end{figure*}

To visualize the agents' behaviors and attitudes, Unity was used as a graphics demonstrator, where all physics simulations are done using JSBSim and sent to Unity for visualization. Fig. \ref{fig:unity_1} shows different episode screenshots for the Dreamer agent pursuing the target. Fig. \ref{fig:unity_1_1} shows the pursuer's point of view while pursuing the evader. Figures \ref{fig:unity_1_2}, \ref{fig:unity_1_3}, and \ref{fig:unity_1_4} show the trajectories of both the evader and the pursuer.

\subsubsection{Robustness Validation Against Perturbations}

Algorithms have been tested against wind-gust perturbations to assess robustness. This validation was done following the setup explained in Section \ref{sec:validation_scenarios}, but in this case, only the Dreamer and TQC agents were studied, given the poor results of SAC in validation under no perturbation conditions.

\begin{figure*}[h]
    \centering
    \begin{subfigure}[b]{0.6\textwidth}
        \centering
        \includegraphics[width=\textwidth]{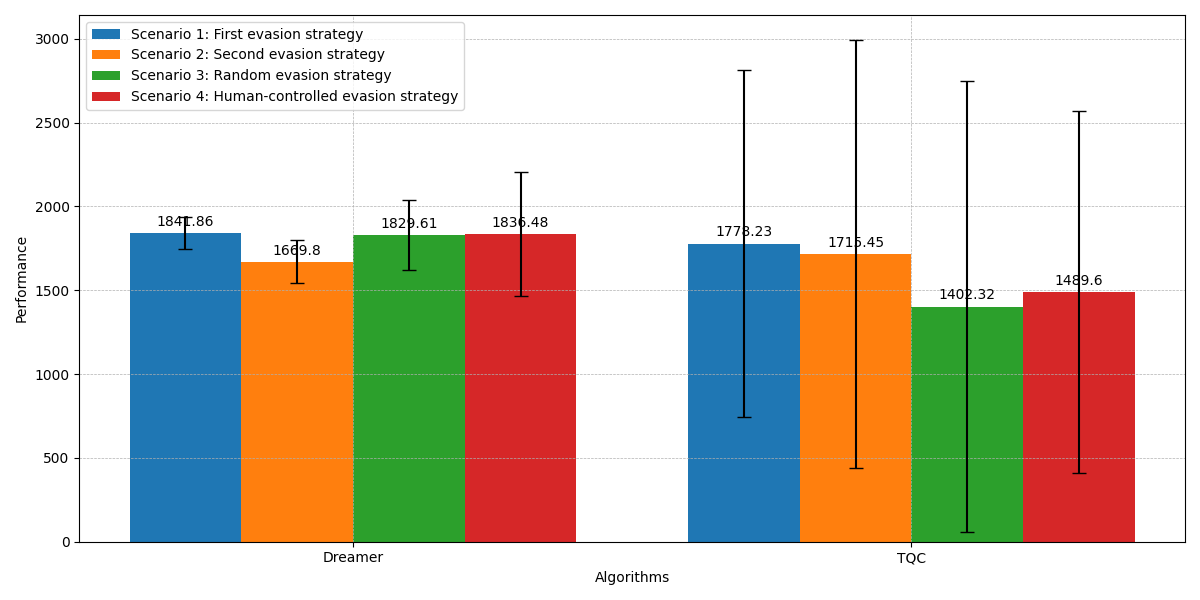}
        \caption{Validation of algorithms across different strategies with wind}
        \label{fig:validation_wind}
    \end{subfigure}
    \begin{subfigure}[b]{0.6\textwidth}
        \centering
        \includegraphics[width=\textwidth]{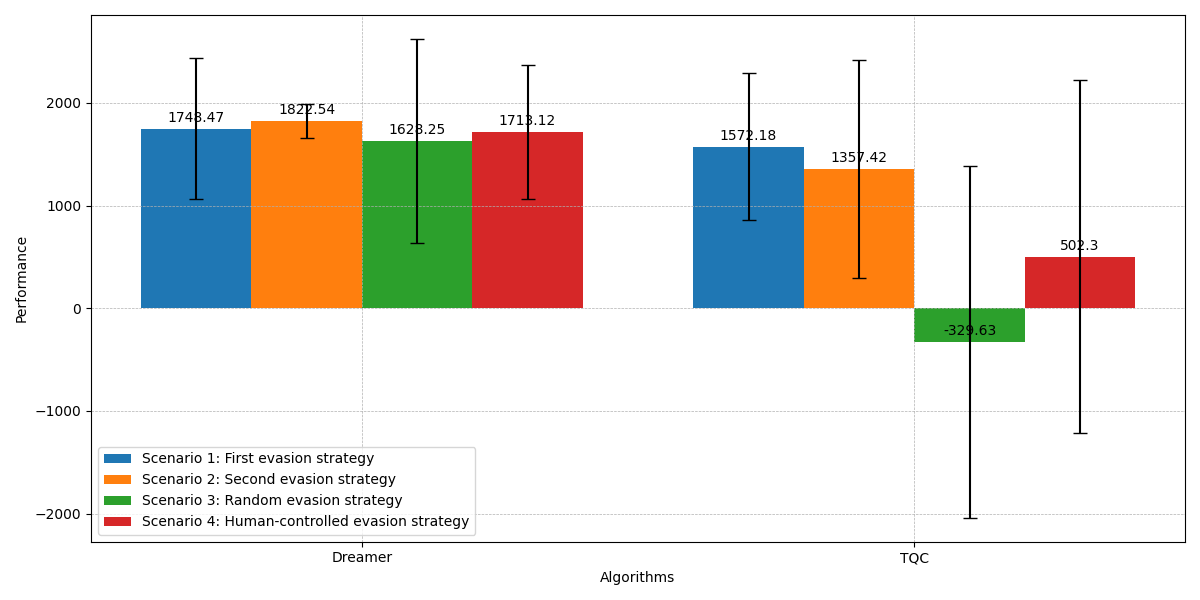}
        \caption{Validation of algorithms across different strategies with noise}
        \label{fig:validation_noise}
    \end{subfigure}
    \caption{Validation studies against perturbations}
    \label{fig:validation_wind_noise}
\end{figure*}

Similar to before, validation was done across all scenarios for each algorithm. Fig. \ref{fig:validation_wind} shows the results of the study against wind for the Dreamer and TQC algorithms across each scenario. These results show that both algorithms have similar performance on average, maintaining results comparable to the cases without wind perturbations. However, whereas Dreamer is stable across episodes, as seen in the standard deviation results, TQC increases its variance considerably compared to validation without wind gusts.

Additionally, the same study was done but evaluating agents' responses against sensor noise instead of wind. This sensor noise is represented as Gaussian noise in the observations, adding a zero-mean Gaussian noise with a 25\% standard deviation for each value in the observation, which translates to high values of noise added in some cases. This is a particularly hard test for reinforcement learning algorithms trained in a fully observable MDP, given that the noise converts the process into a POMDP (Partially Observable Markov Decision Process) \cite{ni2021recurrent}. Fig. \ref{fig:validation_noise} shows the results of this study, again showcasing that the Dreamer agent seems to be more robust against perturbations, being able to reach the target for all evasion strategies, although with reduced performance compared to before. The TQC agent's performance is reduced compared to the other studies, especially for random evasion and human-controlled evasion, being unable to capture the evader when it uses a random evasion strategy. The latent state space of Dreamer, which is composed of current observation and recurrent observations, allows it to handle POMDP.

Finally, both test perturbations were combined to assess the agent in these particular situations. Table \ref{table:comparison_all_results} provides a summary of all the results obtained during the studies.

Given these results, we can assume Dreamer is a more robust algorithm than TQC for these particular scenarios, showcasing stable behavior under perturbations and generalizing to other target behaviors.

\begin{table*}[h]
    \centering
    \caption{Summary of validation results}
    \begin{adjustbox}{max width=\textwidth}
    \begin{tabular}{lcccccccc}
        \toprule
        \textbf{Algorithm} & \multicolumn{2}{c}{\textbf{Validation}} & \multicolumn{2}{c}{\textbf{Wind Gusts}} & \multicolumn{2}{c}{\textbf{Sensor Noise}} & \multicolumn{2}{c}{\textbf{Sensor Noise + Wind Gusts}} \\
        \cmidrule(r){2-3} \cmidrule(r){4-5} \cmidrule(r){6-7} \cmidrule(r){8-9}
        & \textbf{Mean} & \textbf{Std Dev} & \textbf{Mean} & \textbf{Std Dev} & \textbf{Mean} & \textbf{Std Dev} & \textbf{Mean} & \textbf{Std Dev} \\
        \midrule
        \textbf{Dreamer} & & & & & & & & \\
        \midrule
        \textit{Scenario 1} & 1869.49 & 70.13 & 1841.86 & 95.16 & 1748.47 & 685.18 & 1794.70 & 350.24 \\
        \textit{Scenario 2} & 1908.97 & 39.90 & 1669.80 & 128.11 & 1822.54 & 164.36 & 1866.96 & 802.50 \\
        \textit{Scenario 3} & 1849.37 & 97.61 & 1829.61 & 208.55 & 1628.25 & 995.03 & 1820.92 & 138.37 \\
        \textit{Scenario 4} & 1875.97 & 85.86 & 1836.48 & 369.60 & 1713.12 & 652.87 & 1745.98 & 897.26 \\
        \midrule
        \textbf{TQC} & & & & & & & & \\
        \midrule
        \textit{Scenario 1} & 1842.97 & 215.67 & 1778.23 & 1033.50 & 1572.18 & 714.57 & 1585.25 & 1036.09 \\
        \textit{Scenario 2} & 1797.72 & 148.59 & 1715.45 & 1276.64 & 1357.42 & 1065.45 & 1407.92 & 962.19 \\
        \textit{Scenario 3} & 1420.10 & 1069.92 & 1402.32 & 1345.67 & -329.63 & 1710.74 & -581.31 & 1738.21 \\
        \textit{Scenario 4} & 1542.77 & 856.04 & 1489.60 & 1081.40 & 502.30 & 1719.25 & 542.50 & 1740.25 \\
        \midrule
        \textbf{SAC} & & & & & & & & \\
        \midrule
        \textit{Scenario 1} & 26.88 & 1986.92 & - & - & - & - & - & - \\
        \textit{Scenario 2} & 35.37 & 2027.17 & - & - & - & - & - & - \\
        \textit{Scenario 3} & -538.16 & 1975.02 & - & - & - & - & - & - \\
        \textit{Scenario 4} & -20.09 & 2016.19 & - & - & - & - & - & - \\
        \bottomrule
    \end{tabular}
    \end{adjustbox}
    \label{table:comparison_all_results}
\end{table*}

\begin{figure*}[h]
    \centering
    \begin{subfigure}[b]{0.45\textwidth}
        \centering
        \includegraphics[width=\textwidth]{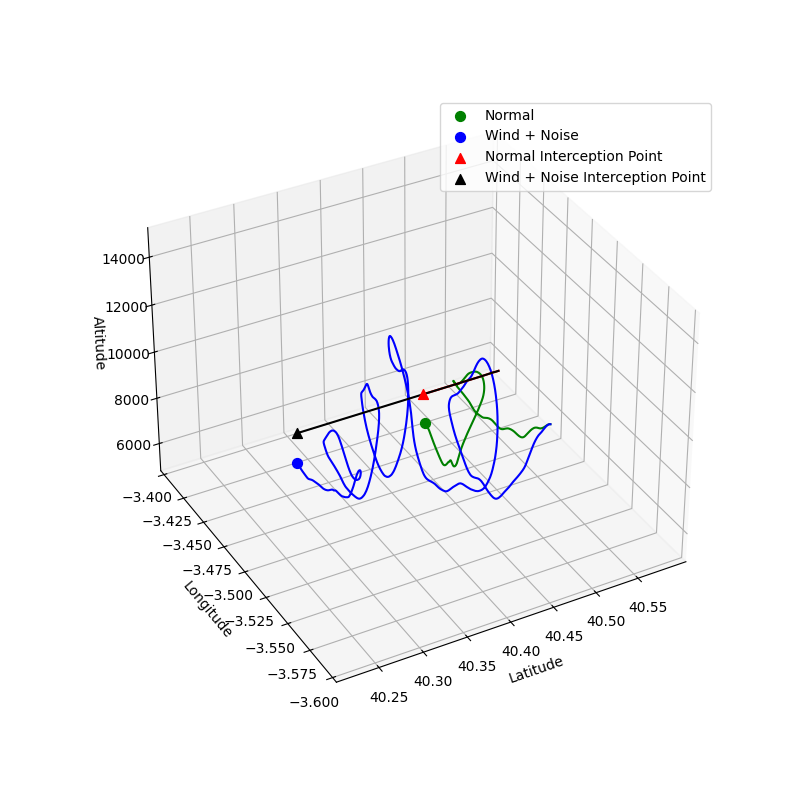}
        \caption{Trajectory comparison until intersection between Dreamer with wind and noise.}
        \label{fig:trajectory_comp_wind_3d}
    \end{subfigure}
    \begin{subfigure}[b]{0.45\textwidth}
        \centering
        \includegraphics[width=\textwidth]{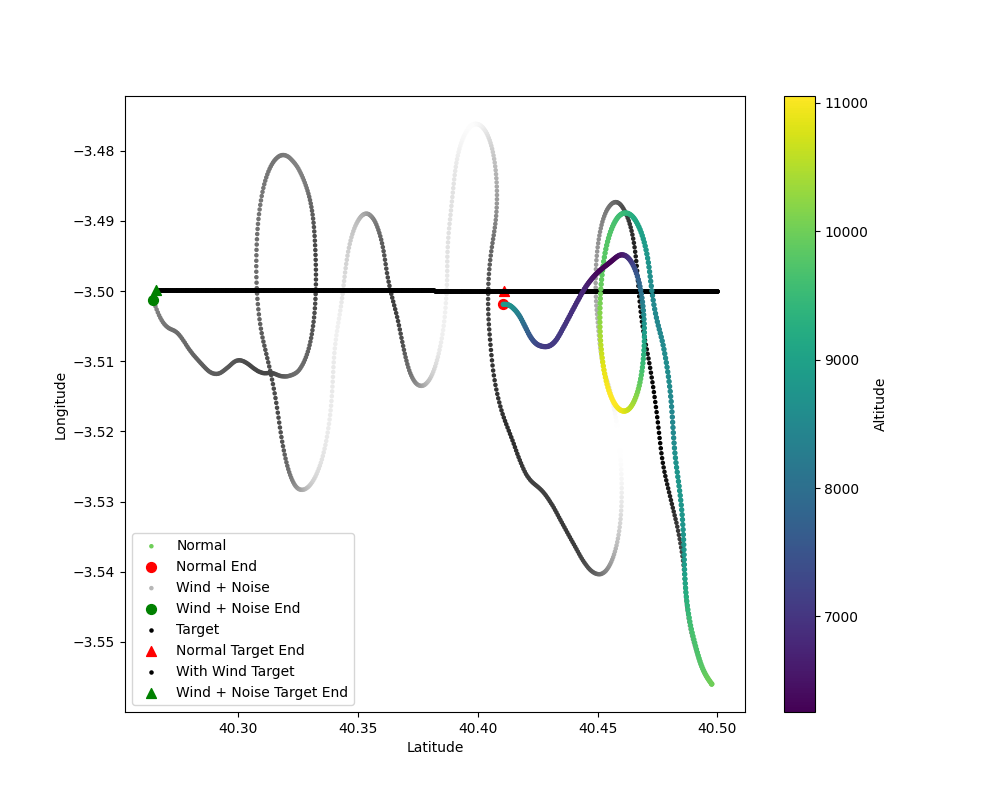}
        \caption{2D trajectory comparison until intersection between Dreamer with wind and noise.}
        \label{fig:trajectory_comp_wind_2d}
    \end{subfigure}
    \caption{Trajectory comparison until intersection between Dreamer with wind and noise.}
    \label{fig:trajectory_comp_wind}
\end{figure*}

In Fig. \ref{fig:trajectory_comp_wind}, a comparison of the trajectories followed by the Dreamer agent for the cases with wind-gust and sensor-noise perturbations and without perturbations is shown. The target is following a fixed line trajectory. The green line shows the trajectory of the Dreamer agent without perturbations, with the intersection point marked in red. The blue line shows the trajectory of the agent with both wind-gust and sensor-noise perturbations, with the intersection point marked in black. Despite the severe wind gusts and sensor noise added, the Dreamer agent is still able to capture the target in a non-seen trajectory during training, showing robustness and generalization capabilities.

\section{Conclusions} \label{conclusions}

This paper presents a comprehensive study on training fixed-wing UAV pursuer agents using Reinforcement Learning (RL) to intercept dynamic evader targets. We compared the performance of model-based and model-free RL algorithms, specifically DreamerV3, TQC, and SAC, under a variety of scenarios, including unseen evasion strategies and perturbations such as wind gusts and sensor noise.

Our findings indicate that DreamerV3 and TQC can efficiently learn the task within the predefined training budget, achieving high performance and stability in the training scenarios. DreamerV3, leveraging its imagination-based training and latent state-space representation, demonstrated superior generalization capabilities and robustness against unseen target behaviors, compared to TQC. While TQC also performed well, it showed greater variability in performance when subjected to different validation scenarios, particularly under perturbations.

In contrast, SAC struggled to achieve the same level of proficiency as DreamerV3 and TQC. Despite benefiting from a pre-training curriculum, SAC exhibited less stability and efficiency, particularly in scenarios involving random and human-controlled evasion strategies.

Validation studies highlighted DreamerV3's robustness, showing consistent performance across all tested evasion strategies and perturbation scenarios. Dreamer's ability to handle partial observability, introduced by sensor noise, underscores its effectiveness in realistic, noisy environments. TQC, while effective in some scenarios, showed significant performance degradation under perturbations, emphasizing the need for more robust model-free approaches.

The robustness tests against wind gusts and sensor noise further solidified DreamerV3's position as a more resilient and adaptive algorithm. It maintained stable performance and efficiently captured the target even under severe perturbations, whereas TQC's performance significantly fluctuated.

In summary, this work underscores the potential of model-based RL, specifically DreamerV3, in complex UAV interception tasks. Dreamer's ability to generalize well beyond training scenarios and maintain robustness under perturbations makes it a promising candidate for real-world applications. Future work could explore further enhancements to model-free algorithms to improve their robustness and efficiency, as well as the application of these RL frameworks to a broader range of autonomous aerial tasks.

The insights gained from this study contribute valuable knowledge to the field of autonomous aerial robotics, highlighting the strengths and limitations of current RL approaches and setting a foundation for future advancements in the domain.

\clearpage


\bibliographystyle{unsrt}
\bibliography{references}  

\end{document}